\documentclass[conference]{IEEEtran}
\IEEEoverridecommandlockouts
\usepackage{cite}
\usepackage{amsmath,amssymb,amsfonts}
\usepackage{algorithmic}
\usepackage{graphicx}
\usepackage{textcomp}
\usepackage{xcolor}
\usepackage{soul}
\usepackage{authblk}
\def\BibTeX{{\rm B\kern-.05em{\sc i\kern-.025em b}\kern-.08em
    T\kern-.1667em\lower.7ex\hbox{E}\kern-.125emX}}
\begin{document}

\title{ProtoTopic: Prototypical Network for Few-Shot Medical Topic Modeling\\}

\author[1]{Martin Licht}
\author[2,3,4]{Sara Ketabi}
\author[2,3,4,5,6,7,8*]{Farzad Khalvati}

\affil[1]{Engineering Science, University of Toronto, Toronto, ON, Canada}
\affil[2]{Neurosciences and Mental Health Research Program, The Hospital for Sick Children, Toronto, Canada}
\affil[3]{Department of Mechanical
and Industrial Engineering, University of Toronto, Toronto, Canada}
\affil[4]{Vector Institute for Artificial Intelligence, Toronto, Canada}
\affil[5]{Institute of
Medical Science, University of Toronto, Toronto, Canada}
\affil[6]{Department of Computer Science, University of Toronto, Toronto, Canada}
\affil[7]{Department of Diagnostic and Interventional Radiology,
The Hospital for Sick Children, Toronto, Canada}
\affil[8]{Department of Medical Imaging, University of Toronto, Toronto,
Canada}
\affil[*]{farzad.khalvati@utoronto.ca}

\maketitle

\begin{abstract}
Topic modeling is a useful tool for analyzing large corpora of written documents, particularly academic papers. Despite a wide variety of proposed topic modeling techniques, these techniques do not perform well when applied to medical texts. This can be due to the low number of documents available for some topics in the healthcare domain. In this paper, we propose ProtoTopic, a prototypical network-based topic model used for topic generation for a set of medical paper abstracts. Prototypical networks are efficient, explainable models that make predictions by computing distances between input datapoints and a set of prototype representations, making them particularly effective in low-data or few-shot learning scenarios. With ProtoTopic, we demonstrate improved topic coherence and diversity compared to two topic modeling baselines used in the literature, demonstrating the ability of our model to generate medically relevant topics even with limited data. 
\end{abstract}


\section{Introduction}

Natural language processing (NLP), as a subset of machine learning (ML), allows for the interpretation and manipulation of language, even with human-level performance. In the healthcare domain, NLP has a wide range of applications, given the recent advances in large language models (LLMs) fine-tuned on clinical tasks. \cite{b1}. NLP is also very useful for topic modeling in clinical settings, which focuses on identifying underlying themes in collections of documents. In particular, topic modeling of medical research papers can be a valuable tool for researchers and clinicians to quickly sort through research papers and find information relevant to their work. Topic models have already achieved strong results in biological text mining \cite{b39, b40, b41, b42, b43, b44, b45}. However, one major challenge of NLP in healthcare is the lack of high-quality training data which most ML algorithms need to train on and is often unavailable in clinical settings. Furthermore, many NLP models lack explainability. In other words, most state-of-the-art models are black boxes which arrive at an output but are unable to explain the reasoning behind that. Finally, there are nomenclature differences which differentiate medical text from general text data. Specific medical terminology as well as formatting and nomenclatures differences between hospitals and institutions present the need for NLP models applied exclusively to medical text data.

To address these challenges, in this work, we propose ProtoTopic, a prototypical network which requires only a small set of samples for training to perform topic modeling on medical research papers. Prototypical networks are explainable deep learning models that work by comparing input data to a set of prototypes, i.e., abstract representations of documents within a dataset learned by the model, to determine which prototype most closely represents a text \cite{b2}. Prototypical networks offer a number of advantages. They address the issue of limited training data per class or group, which is a common feature of clinical datasets, known as few-shot learning, by learning and performing a task based on a small number of data points within each group. To that end, a topic model developed based on a prototypical network would be able to learn topics from just a few documents, making it valuable for use in healthcare systems, where training data can be sparse. Furthermore, it can improve explainability by showing which representative cases most closely describe the text that is being analyzed \cite{b3}. 

To the best of our knowledge, this is the first study that develops a prototypical network for the topic modeling task, taking medical abstracts as inputs and clustering them into distinct topics through comparison with a set of prototype representations. To assess the efficiency of the proposed approach, we compared its performance to that of two state-of-the-art topic modeling algorithms, namely Latent Dirichlet Analysis (LDA) \cite{b4} and BERTopic \cite{b8}. Our contributions in 
this study can be summarized as: 

\begin{itemize}
    \item Developing a prototypical network for topic modeling on medical abstracts.
    \item Comparing the performance of the proposed model against two baseline models, namely LDA and BERTopic. 
    \item Analyzing the effect of the number of topics on the overall model performance. 
\end{itemize}

\section{Literature Review}

\subsection{Topic Modeling}
Probabilistic topic models have been widely explored in the form of Latent Dirichlet Allocation (LDA) \cite{b4}. This model employs the Bag of Words (BoW) assumption where the order of words in a given document is disregarded and only the frequency of words is considered relevant. 

Neural topic models, on the other hand, have been more recently explored and are able to capture word context using text embeddings. LDA2VEC \cite{b5} uses Word2Vec \cite{b6} embeddings alongside LDA to capture word context by analyzing a window of words and learning to predict new words given the context. Embedded Topic Model (ETM) \cite{b7} also uses LDA but uses the Continuous Bag of Words (CBOW) embeddings rather than Skip-gram embeddings. BERTopic \cite{b8} is a topic model based on the Bidirectional Encoder Representations from Transformers (BERT) \cite{b9} embeddings. This model generates embeddings over the entire text, performs dimensionality reduction, and finally clusters the reduced embeddings to generate topics. BERTopic most closely matches our approach to topic modeling.

\subsection{Few-shot Learning}
Few-shot Learning (FSL) strategies can largely be divided into two categories based on their approach: optimization- based and metric-based. Optimization-based (or parameter updating-based) approaches aim to predict the updating model parameters based on the limited data available. These approaches often rely on meta-learning, a form of learning which focuses on optimizing the learning process itself such that a model can learn patterns based on very few examples. Ravi and Larochelle \cite{b10} developed a Long Short-Term Memory (LSTM)-based meta learner which aimed to learn efficient parameter updating rules and a general initialization of parameters to allow for quick convergence. Finn et al. \cite{b11} developed a model-agnostic meta-learning (MAML) algorithm which aims to produce a parameter weight initialization which allows for efficient training for any gradient-based ML model.

Metric-based approaches focus on learning a generalizable metric function which can be used to compute the similarities between instances across tasks. Koch et al. \cite{b12} developed a Siamese network which uses convolutional neural networks to extract information from an image and then computes a metric determining the image similarity with other images. The weights of this network can be efficiently learned across limited training samples and then be generalized to classes associated with very few examples to analyze (the paper focused on the one-shot learning scenario). Vinyals et al. \cite{b13} proposed Matching Networks, a model which uses memory- augmented neural networks \cite{b14,b15} comprised of an external memory and an attention mechanism applied to access the memory. The matching network learns separate embedding functions for support and query sets and is then able to use these embeddings with the stored memory to obtain useful classifications for new examples. The support set is the set of datapoints used for training purposes and the query set is the set of datapoints used to evaluate the performance on the task. The matching network uses the meta-learning and memory augmentation approach of Santoro et al. \cite{b16} but applies it to image data instead of sequential data using an LSTM. Sung et al. \cite{b17} developed the Relation Network which learns a deep distance metric during training and can then classify new images by calculating relation scores between query images and just a few examples of new classes.

\subsection{Prototypical Networks}
Prototypical networks are another approach used for FSL based on a metric-based strategy. They were first introduced by Snell et al.~\cite{b2} as a tool for FSL and were found to be extremely effective by addressing the key issue of overfitting in scenarios with limited or no labeled training data. Prototypical networks were first proposed for image classification tasks \cite{b18}. Many prototypical networks have since been developed to improve the few-shot capabilities for computer vision applications \cite{b19,b20,b21,b22,b23,b24,b25}. The concept of prototypical networks was also implemented strongly in the field of NLP by extracting latent representations of the text which could then be compared to a set of prototypes. The idea was adapted for sequential text classification with the ProSeNet model \cite{b26}. This model features LSTMs in a recurrent sequence encoder which generates text representation. This representation is then compared to the prototypes and their similarity is used as the sole input to a fully connected layer which outputs the classification task probabilities. In ProtoryNet \cite{b27}, the prototypes were formed from sentences instead of whole documents, using the pretrained DistilBERT model \cite{b28} to generate sentence embeddings. ProtoSeq \cite{b29} is a sequential prototypical network which incorporates an LSTM as well as a Convolutional Neural Network (CNN) to perform few-shot emotion recognition in conversation data. Plucinski et. al \cite{b30} introduced a prototype-based CNN which uses phrases as prototypes for a sentiment classification model. ProtoAttend \cite{b31} demonstrated the capabilities of a prototypical network combined with an attention mechanism. Finally, Proto-lm \cite{b32} combines the impressive capabilities of LLMs, such as BERT \cite{b9}, with a prototypical layer for text classification tasks.

\section{Materials and Methods}

\subsection{Data and Data Preprocessing}
The dataset used in this study is PubMed200k RCT \cite{b33}, which is an open-sourced collection of 200,000 abstracts of randomized control trials (RCT) from the PubMed database.  The dataset consists of 2.3 million sentences, each labeled based on their role in the abstract as one of the following: background, object, method, result, or conclusion. However, as the goal of this was to perform topic modeling on the data, we did not require these labels and during data processing, the labels were completely ignored. We ensured topic diversity by using a large enough dataset and also used a publicly available dataset, thereby making our experiments reproducible.   

To preprocess the text, we performed several steps including:

\begin{itemize}
\item Removing non-alphabet characters, e.g., numbers and dates.
\item Converting text to lower-case
\item Text Tokenization (which is performed by the text encoder)
\item Removing high-frequency words, e.g., “the”, “and”, “of”, etc.
\end{itemize}

\subsection{Methodology}

\begin{figure}
\centerline{\includegraphics[scale=0.3]{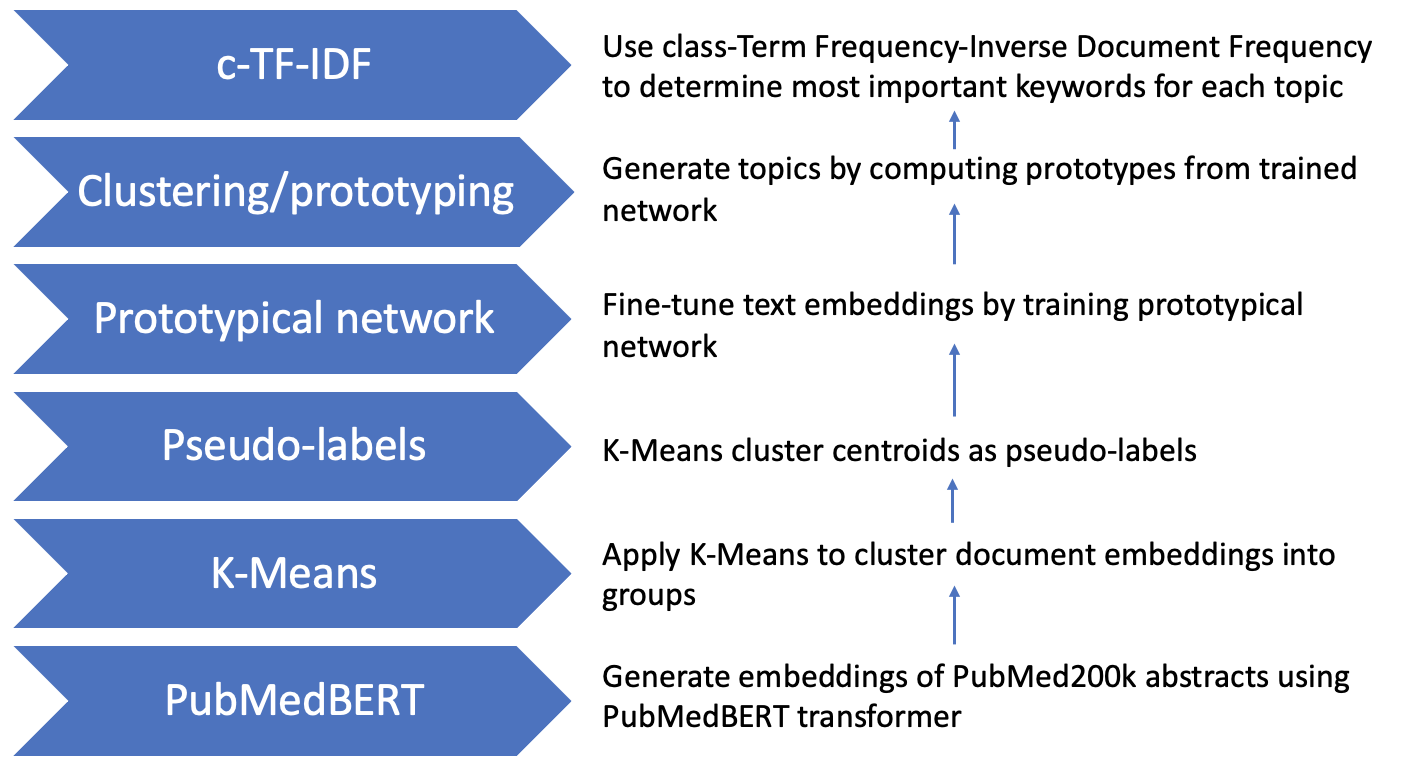}}
\caption{ProtoTopic Training Pipeline}
\label{fig1}
\end{figure}

 To develop ProtoTopic, we performed several sequential steps, as demonstrated in Fig. \ref{fig1}. These steps include generating embeddings for the abstracts using two language transformers: PubMedBERT \cite{b34} and ``all-MiniLM-L6-v2", clustering the embeddings using K-means, using the K-means-extracted centroids as pseudo-labels, applying the pseudo-labels to train a prototypical network, getting the prototypical network to cluster the abstracts into distinct topics, and finally, using class-based Term Frequency-Inverse Document Frequency (TF-IDF) to extract representative words for each topic. These steps are explained in detail below. 

The first step for developing any ML model is transforming the input data into some numerical form that can be processed later by the model. In ProtoTopic, this was performed by using two separate attention-based transformers to create text embeddings. The first one was PubMedBERT \cite{b34}, a variant of the BERT \cite{b9} transformer trained on medical papers, to capture domain-specific medical embeddings. Specifically, each abstract was converted into a 768-dimensional vector to encode the semantic information within the text. The second transformer used was all-MiniLM-L6-v2, a general-purpose transformer, converting the abstracts into 384-dimensional vectors. 

Following embedding generation, we applied K-means to the embeddings to cluster them into distinct groups. K-means is an unsupervised algorithm that partitions input data into a predefined number of clusters by minimizing the variance within each cluster through iterative assignment and centroid updating. The output of this step is a set of centroids where each document is assigned to a single centroid, specifying which cluster the document belongs to. These centroids were used as pseudo-labels for training the proposed prototypical network.

Therefore, the next step was training our prototypical network using the K-means-extracted pseudo-labels. A schematic representation of this model is provided in Fig. \ref{fig1}. As in \cite{b2}, we have a small support set of N labeled examples \(S = \{(\textbf{x}_1,y_1),...,(\textbf{x}_N,y_N)\}\). We obtain these support examples from our K-means clustering where the \(\textbf{x}_i\)'s are our abstract representations and \(y_i\in\{1,...,K\}\) are the corresponding class labels. \(S_k\) is the set of all support examples labeled with class \(k\). In each episode then, training was performed by computing the prototypes \(\textbf{c}_k\) by taking the mean of the support set \(S_k\) after applying our embedding function \(f_\phi\) to our abstracts \(\textbf{x}_i\) (Equation~\ref{eq1}):

\begin{equation}
    \textbf{c}_k = \frac{1}{|S_k|} \sum_{(\textbf{x}_i,y_i)\in S_k}f_\phi(\textbf{x}_i)
\label{eq1}
\end{equation}

\begin{figure}
\centerline{\includegraphics[scale=0.4]{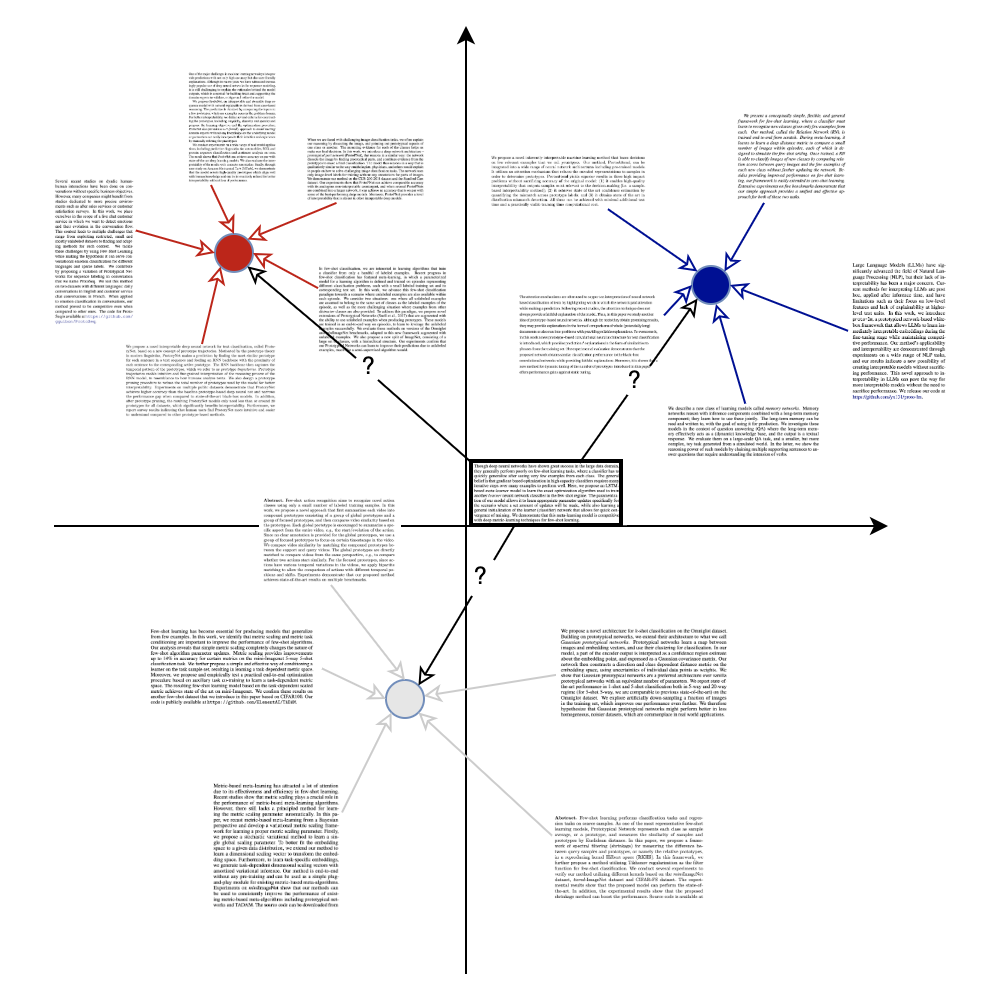}}
\caption{A Schematic Overview of the Proposed ProtoTopic Model: Abstracts are shown in a 2D representation of the PubMedBERT/all-MiniLM-L6-v2 embedding space. Each of the 3 topics has 5 abstract embeddings in its support set. The prototypes (red, blue and grey points) are computed by taking the mean of the support set embeddings, and the query point (abstract in text box) is then compared to the prototypes to predict which topic it corresponds to.}
\label{fig4}
\end{figure}

We then determined the classes of the query set, referring to the datapoints for which we intend to make predictions, by finding the closest prototype to each point embedding in this set (see Fig. \ref{fig4}). This allowed us to define a probability that a query point belonged to a given class using the softmax of the distances \(d\) between query points \(\textbf{x}\) and prototypes, as shown in Equation~\ref{eq2} (\(k'\) refers to all classes including \(k\)).

\begin{equation}
    p_\phi (y=k|\textbf{x})=\frac{exp(-d(f_\phi (\textbf{x}),c_k))}{\sum_{k'}exp(-d(f_\phi (\textbf{x}),c_{k'}))}
\label{eq2}
\end{equation}

Subsequently, we minimized the loss \(J(\phi)=-\log p_\phi(y=k|\textbf{x)}\) between assigned classes and initial pseudo-labels. During this process, the PubMedBERT/all-MiniLM-L6-v2 transformers were iteratively fine-tuned to improve the quality of text embeddings using the computed prototype representations from different steps.

After clustering the data into distinct groups using our prototypical network, we then needed a method to extract representative keywords describing a given topic. Consequently, we applied a class-based TF-IDF (c-TF-IDF) method developed by authors of BERTopic \cite{b8}. This algorithm amended the TF-IDF model proposed in \cite{b35} to improve its use for class-based algorithms (such as BERTopic and ProtoTopic). This was achieved by redefining word frequency, from the proportion of groups in which a word appears to the percentage of the word’s occurrences across all groups. This greatly improved performance in the case of BERTopic and ProtoTopic, as there were not many groups, but each group is composed of up to thousands of documents. In the case of TF-IDF, there would be no difference between a group containing a word a single time or it containing the word a thousand times. For this reason, the c-TF-IDF method performed much better in extracting representative keywords for topic modeling, and this is why the algorithm was chosen for ProtoTopic.

In order to ensure that the results of ProtoTopic on the PubMed200k were not simply a result of the PubMedBERT transformer being trained on biomedical text data, two versions of the model were fine-tuned. The main ProtoTopic model was built as described above, but the second version was created by replacing the PubMedBERT embeddings with all-MiniLM-L6-v2 embeddings. This is a general-purpose state-of-the-art sentence transformer generating 384-dimensional embeddings for input text.

We trained the proposed framework trained using 3 different numbers of topics: 25, 50 and 100. For all experiments, the number of topics was set to the number of clusters initiated by the K-means algorithm. The ADAM optimizer \cite{b36} was applied with a learning rate of 0.00005, 50 episodes per epoch, and a total of 10 epochs. Each episode involved 5 groups, each associated with 5 support set examples and 5 query points. In the pre-trained PubMedBERT and all-MiniLM-L6-v2 transformers, all layers were frozen except for the last two. The model was trained using a T4 GPU with 15GB RAM on Google Colab.

\subsection{Evaluation Setting}

To evaluate the performance of the proposed framework in generating useful topics for medical texts, we trained two baseline models on the dataset, namely LDA and BERTopic. LDA was chosen as one of the baselines as it is the most used topic modeling algorithm and is extensively studied in the literature. It also provides a baseline model which does not employ text embeddings, setting it apart from our other baselines and the proposed model. BERTopic leverages attention-based transformer embeddings to produce contextualized text representations. This model was chosen as the second baseline as it is a strongly performing neural topic model, and we aimed to analyze the impact of the semantic embeddings on the topic modeling performance. 

The models were evaluated based on two metrics, topic coherence and diversity. Although the methods for measuring the coherence vary, some form of coherence score is standard for measuring the performance of topic models. The coherence function takes the corpus (set of documents analyzed by the topic model), the vocabulary, and the top N words generated by each topic and then outputs a score (typically between 0 and 1) based on how well the words describing each topic cohere to one another. For the purposes of our model, we used the coherence score \(C_V\) which is a commonly used coherence metric and was shown in \cite{b37} to correlate the strongest with human ratings of coherence. \(C_V\) works by analyzing the co-occurrence of topic keywords within the corpus to determine how semantically related they are. This measure serves as a tool to compare the performance of the baseline topic models to our proposed ProtoTopic framework. Topic diversity was measured as in \cite{b7} by extracting the top 25 keywords for each topic and then calculating the percentage of unique words in the set of keywords across all topics. Achieving a high topic coherence score and high topic diversity score indicates that the model can generate a diverse set of topics with coherent keywords while being very descriptive and avoiding repetitiveness.

\section{Results}

\subsection{Quantitative Results}
The coherence score and topic diversity metrics were calculated based on the topics generated by LDA, BERTopic, ProtoTopic (with all-MiniLM-L6-v2) and ProtoTopic (with PubMedBERT) for 25 topics (Table \ref{tab1}), 50 topics (Table \ref{tab2}) and 100 topics (Table \ref{tab3}). The results show that for 25 topics, ProtoTopic with PubMedBERT achieves the highest coherence score, ProtoTopic with all-MiniLM-L6-v2 achieves the second-highest score, BERTopic achieves the next-highest score, and LDA achieves the lowest score. For 50 and 100 topics, BERTopic and LDA once again have the second-lowest and lowest scores, respectively, and ProtoTopic with PubMedBERT and all-MiniLM-L6-v2 have the highest scores, with all-MiniLM-L6-v2 barely outscoring PubMedBERT. For topic diversity, ProtoTopic with PubMedBERT outscores ProtoTopic with all-MiniLM-L6-v2 and repeatedly, BERTopic and LDA have the second-lowest and lowest scores, respectively. The topic coherence score increases for each model with the number of topics. i.e., coherence is higher with 100 topics than with 50, and higher with 50 topics than with 25. The same trend holds for topic diversity, except for ProtoTopic. The topic diversity decreases as the number of topics increases for ProtoTopic with both PubMedBERT and all-MiniLM-L6-v2. 

A statistical test (T-test) was also performed to determine the significance of the difference between the performance of the baseline and proposed models. To that end,  ProtoTopic (with PubMedBERT embeddings) and BERTopic  (the highest performing baseline) were evaluated 7 times each with 25 topics, and the coherence and diversity scores were calculated. We then tested the null hypothesis: The mean coherence and diversity scores for ProtoTopic and BERTopic are the same. This analysis yielded a p-value of less than 0.00001 for both coherence and diversity scores. As a result, we have shown that ProtoTopic significantly outperforms BERTopic based coherence and diversity metrics.

\begin{table}[h!]
\caption{Coherence score and topic diversity for ProtoTopic and baseline models evaluated with 25 topics}
\begin{center}
\begin{tabular}{|c|c|c|}
\hline
\textbf{25 topics} & \textbf{Coherence Score} & \textbf{Topic Diversity} \\
\hline
LDA & 0.4910 & 40.8\% \\
\hline
BERTopic & 0.5137 & 49.6\% \\
\hline
ProtoTopic (all-MiniLM-L6-v2) & 0.5396 & 84.5\% \\
\hline
ProtoTopic (PubMedBERT) & \textbf{0.5754} & \textbf{86.1}\% \\
\hline
\end{tabular}
\label{tab1}
\end{center}
\end{table}

\begin{table}[h!]
\caption{Coherence score and topic diversity for ProtoTopic and baseline models evaluated with 50 topics}
\begin{center}
\begin{tabular}{|c|c|c|}
\hline
\textbf{50 topics} & \textbf{Coherence Score} & \textbf{Topic Diversity} \\
\hline
LDA & 0.5017 & 43.8\% \\
\hline
BERTopic & 0.5394 & 54.5\% \\
\hline
ProtoTopic (all-MiniLM-L6-v2) & \textbf{0.6789} & 73.5\% \\
\hline
ProtoTopic (PubMedBERT) & 0.6734 & \textbf{75.9\%} \\
\hline
\end{tabular}
\label{tab2}
\end{center}
\end{table}

\begin{table}[h!]
\caption{Coherence score and topic diversity for ProtoTopic and baseline models evaluated with 100 topics}
\begin{center}
\begin{tabular}{|c|c|c|}
\hline
\textbf{100 topics} & \textbf{Coherence Score} & \textbf{Topic Diversity} \\
\hline
LDA & 0.5090 & 55.6\% \\
\hline
BERTopic & 0.6173 & 58.0\% \\
\hline
ProtoTopic (all-MiniLM-L6-v2) & \textbf{0.7173} & 58.6\% \\
\hline
ProtoTopic (PubMedBERT) & 0.7117 & \textbf{61.2\%} \\
\hline
\end{tabular}
\label{tab3}
\end{center}
\end{table}

\subsection{Qualitative Results}
\begin{figure}[h!]
\centerline{\includegraphics[scale=0.6]{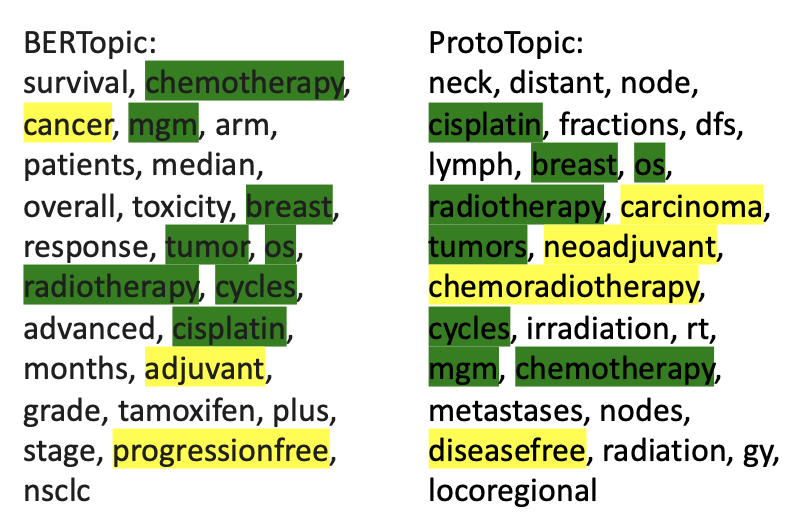}}
\caption{Similarities between BERTopic (left) and ProtoTopic (right) keywords for "cancer" topic.}
\label{fig2}
\end{figure}

Ultimately, the goal of the topic model is to provide an easily interpretable collection of topic keywords for the user which allows them to gain an understanding of the topic held within a corpus. This means that the qualitative behaviour of the model is equally, if not more, important compared to the quantitative results. A topic model output with high coherence score which cannot be understood well by the user is not a good one. In order to fully understand the quality of the topics extracted from a topic model, it is necessary to qualitatively inspect the output of the algorithm. To that end, we analyzed the generated topic keywords to determine the differences and similarities between the output of ProtoTopic (with PubMedBERT) and the baselines. Fig. \ref{fig2} shows the comparison between 2 topics from BERTopic and ProtoTopic that correspond to “cancer”. We have highlighted words that are the same (green) or similar (yellow) across the 2 topics. 

The differences between these outputs were also analyzed. It was found that BERTopic had topics that contained very general keywords related to all documents across the corpus. This was not found to be the case in ProtoTopic, where every topic was specific and only related to a subset of all documents. This can be seen in Fig. 4, which shows the BERTopic keywords for a topic, where the keywords are mostly overarching and general, relevant to many papers. In contrast, ProtoTopic generated very specific keywords, such as those related to common lower body injuries (see Fig. 4) and avoids generating very general topics that are not highly useful.

\begin{figure}[h!]
\centerline{\includegraphics[scale=0.4]{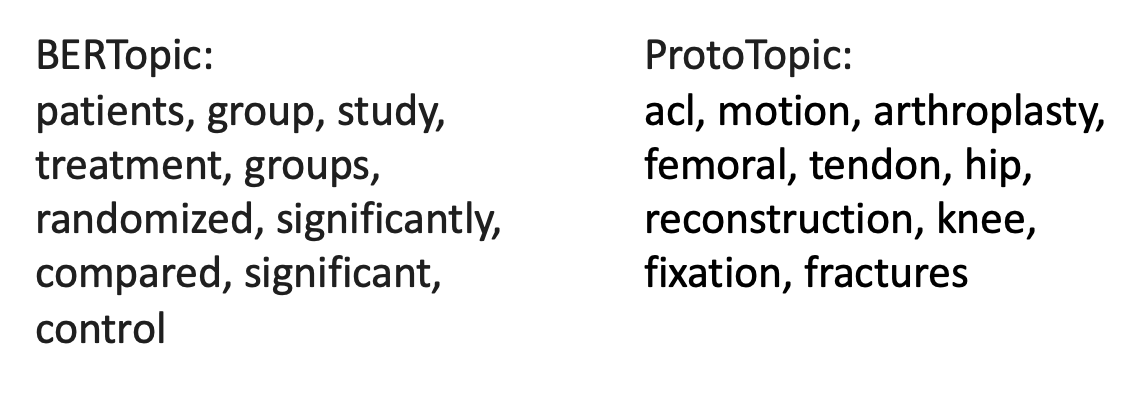}}
\caption{A comparison between the specificity of a set of keywords generated by BERTopic (left) and ProtoTopic (right).}
\label{fig3}
\end{figure}

\section{Discussion}

In this work, we proposed a prototypical network framework, ProtoTopic, for topic modeling on medical abstracts using a limited number of training datapoints per topic. Our results indicate that ProtoTopic achieves improved performance on this task compared to both LDA and BERTopic, as the baselines, on the PubMed200k dataset. This is demonstrated by higher coherence score and topic diversity scores across all topic numbers queried. Therefore, ProtoTopic can be used to generate highly coherent and diverse topics for a corpus of medical research paper abstracts. 

According to our qualitative results, BERTopic generates some topics that contain very general keywords which do not seem to be specific to individual documents (Fig. \ref{fig3}). These keywords do not provide any useful information and cannot be used to differentiate between topics in a given corpus. This can also be seen in Fig. \ref{fig2}, where BERTopic generates uninformative keywords, such as “patients”, “median”, “overall”, and “plus”, which would likely be common across many different topics. Such words are not seen among the topic keywords generated by the ProtoTopic model (see Fig. 3 and Fig. 4), demonstrating the high topic diversity achieved by ProtoTopic. 

An interesting trend observed in the data is that topic coherence and topic diversity increase for all models as the number of topics increases, except for ProtoTopic’s topic diversity. The topic coherence is expected to increase as the number of topics increases because more topics leads to individual topics becoming more specific, and as a result, the topic keywords can be more closely related. This trend is seen for all models. One might expect the topic diversity to decrease as the number of topics increases. The reasoning behind this is that a higher number of topics leads to more total keywords, which could decrease the probability that a keyword becomes unique. However, the opposite trend is seen for the baseline models, where the topic diversity increases as the number of topics increases. One possible explanation for this is that when the number of topics is low, each topic must accommodate a very large number of documents. If the documents are sufficiently diverse, then the keywords for any given topic must be very general to properly describe a wide range of document topics. As the number of topics increases, the topics would then become more specific and would shed the overly general keywords, resulting in more unique words and higher topic diversity, despite the total number of keywords increasing. ProtoTopic, on the other hand, sees a sharp decrease in topic diversity as the number of topics increases. One possible explanation for this could be the fact that the diversity starts at a very high value (86.1\% with 25 topics) because the model is very good at avoiding words which are common across many topics. However, as the number of topics increases, the total number of keywords increases, resulting in more overlapping words and a lower topic diversity as mentioned above. This effect is visualized in Fig. 5. We initially have our two models with low topic diversity (BERTopic) and high topic diversity (ProtoTopic) at 10 topics. As the number of topics increases, BERTopic's topics become more specific, resulting in less overlap in the topics. In contrast, ProtoTopic stars with high topic diversity as it generates very specific topics. As the number increases, the topic diversity decreases as there is more overlap between the topics.

\begin{figure}[h!]
\centerline{\includegraphics[scale=0.5]{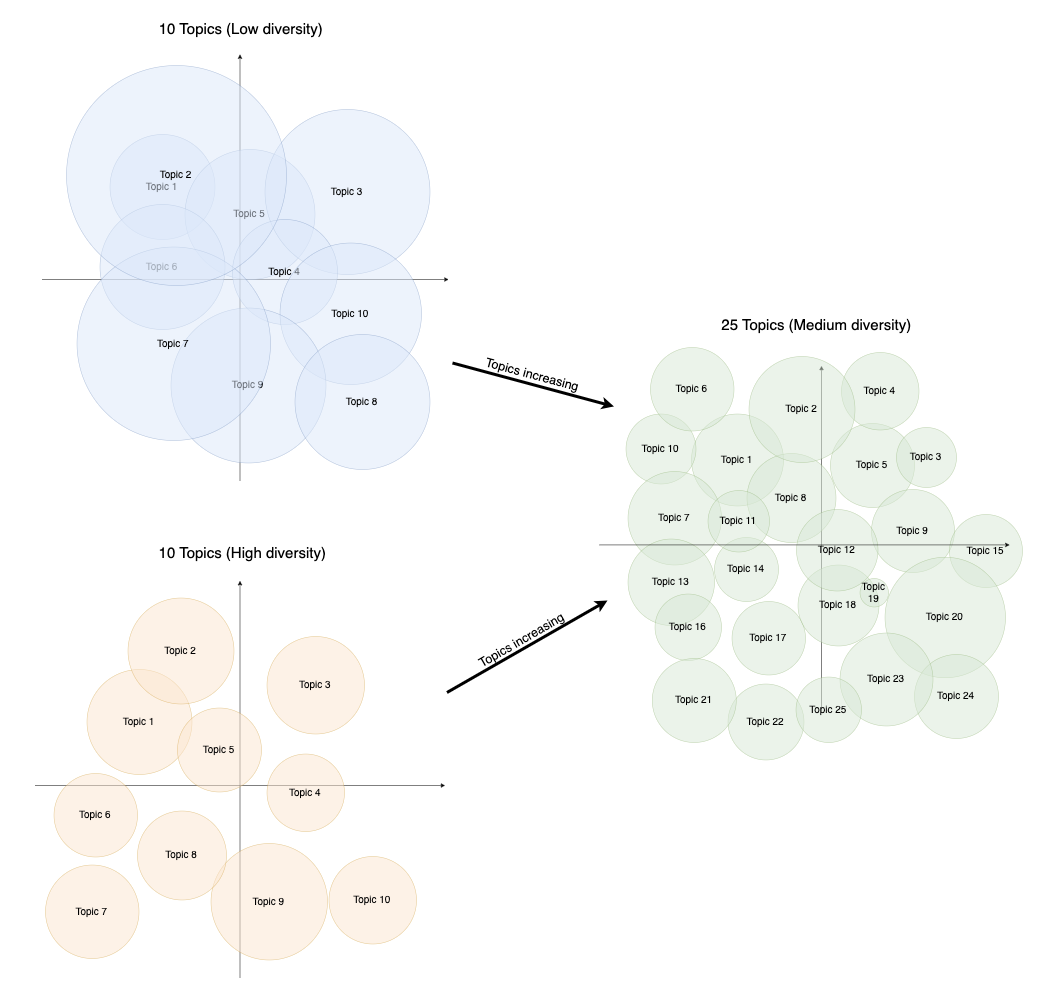}}
\caption{A Schematic Overview of the Effect of Topic Number on Topic Diversity: Topics are shown in a 2D representation of the embedding space. We initially see 10 topics with low diversity (top left) due to general topics and 10 topics with high diversity (bottom left) and less topic overlap. As the topic number increases, the topic diversity converges.}
\label{fig5}
\end{figure}

It is noteworthy that the increased coherence and topic diversity observed when applying ProtoTopic to the PubMed200k dataset does not seem to simply be a result of the increased specificity of the PubMedBERT embeddings, as ProtoTopic outperforms the baselines even when using a general transformer, i.e., all-MiniLM-L6-v2, instead of PubMedBERT. The coherence score is somewhat higher for ProtoTopic with PubMedBERT embeddings at 25 topics. However, the achieved scores are highly similar at 50 and 100 topics when using PubMedBERT embeddings and all-MiniLM-L6-v2 embeddings in ProtoTopic. Moreover, ProtoTopic with PubMedBERT outperforms all-MiniLM-L6-v2 in terms of topic diversity across all topic numbers, and all-MiniLM-L6-v2 surpasses both baselines. This indicates that the high performance of ProtoTopic is a result of the effective architecture of the model, rather than simply the embedding model choice.

There are some areas worth investigating in future work. The current loss function is the one introduced in \cite{b2} for prototypical network training. In more recent works \cite{b3,b32}, additional loss terms are introduced to ensure tight clusters and prototypes which are spaced out. Furthermore, different clustering techniques other than K-means can also be explored. For instance, BERTopic employs HDBSCAN \cite{b38}, which could be used in future work. The effect of dimensionality reduction could also be explored since the PubMedBERT transformer generates very high-dimensional (768-dimensional) embeddings. Reducing the dimension of these embeddings (a strategy used by BERTopic) could lead to improved performance if it does not remove important semantic information. Finally, the fidelity of the topics and topic keywords generated by the model could be evaluated through a user study by a clinician. This could evaluate whether the generated topics and keywords are indeed more specific and interpretable than those generated by the baseline models.

\section{Conclusion}

In this paper, we proposed a prototypical network for the task of topic modeling on medical research literature. Our model, ProtoTopic, achieved superior topic coherence and diversity compared to topic model baselines, LDA and BERTopic. We also qualitatively demonstrated the generation of medically relevant, interpretable topics and corresponding keywords, allowing for quick and efficient understanding of the topics present in the dataset. The findings of this research pave the way for investigating the few-shot performance of prototypical networks on the task of topic modeling, improving the ability of ML models to generate high-quality topics even with limited data on certain topics.

\end{document}